# BPCMont: Business Process Change Management Ontology

Muhammad Fahad

DISP Lab (http://www.disp-lab.fr/), Université Lumiere Lyon 2, France

muhammad.fahad@univ-lyon2.fr

*Abstract*—Change management for evolving collaborative business process development is crucial when the business logic, transections and workflow change due to changes in business strategies or organizational and technical environment. During the change implementation, business processes are analyzed and improved ensuring that they capture the proposed change and they do not contain any undesired functionalities or change side-effects. This paper presents *Business Process Change Management approach* for the efficient and effective implementation of change in the business process. The key technology behind our approach is our proposed *Business Process Change Management Ontology (BPCMont)* which is the main contribution of this paper. *BPCMont*, as a formalized change specification, helps to revert BP into a consistent state in case of system crash, intermediate conflicting stage or unauthorized change done, aid in change traceability in the new and old versions of business processes, change effects can be seen and estimated effectively, ease for Stakeholders to validate and verify change implementation, etc.

## I. INTRODUCTION

In recent years, *Business Process Management (BPM)* emerged as a challenging field that focuses on the set of activities performed by an enterprise to manage and improve its capacities by boosting their business processes. A *Business Process (BP)* is defined as *"a process is an ordering of activities with a beginning and end: it has inputs (in terms of resources, materials, and information) and a specified output (the results it produces)"* [1]. Through business processes, an enterprise can realize and deliver value from its implemented process-based assets. One of the goals of our ongoing research project named *Future Internet Technologies for MANufacturing industries (FITMAN)* is to develop collaborative business process use case trails in the *Smart*, *Digital* and *Virtual* Factories by assessing and testing FI-WARE Generic Enablers [2].

There are many business process modeling notations and languages in the research literature. Some of them are Business Process Modeling Notation (BPMN) [3], Workflow [4], Petri Net [5], Unified Modeling Language (UML) [6] and Business Process Modeling Language (BPML) [7]. Each language provides different notions, syntax, and complexity for modeling business processes. The Business Process Model and Notation is a standard for building business processes developed by Object Management Group (OMG) and is widely used [3]. It provides a graphical notation and equivalent XML for the business process executable language constructs. Based on the BPMN 2.0 specification, researchers have also developed the BPMN ontology [7] to provide quick and unambiguous understanding of its formal specification.

Mostly a business process is modelled in the collaborative environment among different enterprises and participants. With the time and need, business logic and strategies evolve leading changes in their business processes. To cope with the evolving nature of enterprises, changes are natural in the collaborative business process design and implementation. "*To improve is to change; to be perfect is to change often*" [8], fits equally well for the evolving business processes. During the change implementation, business processes are analyzed and improved ensuring that they capture the proposed change and they do not contain any undesired functionalities or change effects. Therefore, it is of immense need to capture changes into a formal specification, systems and structures. In this paper, we are proposing an ontology for the *Business Process Change Management (BPCM)* so that different stakeholders can communicate, refine, and improve their BPs with their evolving business and environment effectively. This ontology (BPCMont) serves as a fundamental building block of our business process change management approach. Following are the motivations behind our proposed BPCM ontology:

*Change Formalization:* Change is formally captured in the BPCM ontology that serves well in the heterogeneous collaborative business process environment where changes can affect different strategies of business and delivered services.

*Revert into a Consistent State:* All the changes and history is maintained inside the BPCM ontology that helps to revert BP to a previous consistent state is the case of ambiguous or conflict occurrence in the collaborative business process update environment or unauthorized user has made a change.

*Change Traceability:* Change is well traceable in the BP in the collaborative environment and among different versions of BP.

*Business Process Recovery:* System crash or damage to BP artifacts, then business process recovery is possible effectively with an ease with the help of BPCM Ontology when all the changes are captured in a formal and semantically organized way.

*Visualization of Change: W*ell formalized Change in the BPCM ontology can help to estimate change effects and potential impacts more evidently, and implications of the change can be well considered as compared to an informal change description.

*Provenance Information*: BPCM ontology also provides information regarding who has made a change for what cause/need and at what time.

*Effective Versioning*: All the versions of BP with the changes as a formal specification serve well for the effective versioning of BPs. It enforces the ability to handle an evolving collaborative business process environment.

*Performance subject to Managed/Unmanaged Change*: Performance of an enterprise is related to the change management due to the fact that strong change management can control/show the capacities of an enterprise in adapting their business, strategies, applications, and also foster the collaborations between enterprises, etc.

*Ease for Stakeholders and participants.* Formal specification helps all the involved stakeholders to observe through the process and proposed changes in a step-by-step manner for the validation and verification of change implementation and to analyze whether the changed BPs actually expose the desired behavior.

This work is much influenced by the evolving schema and ontology domains. These domains have received great attention in the last years. The evolution of *Schema* deals with the study of changes in the schema of populated database without losing its data [9]. Similarly, the evolution of *Ontology* involves the study of changes in the ontology description (*TBox*) and their instances (*ABox*) [10]. There are some works for the evolving business process management, but this domain receives only theoretical proposals and models for the change management yet. Therefore, we took this initiative to design BPCM ontology that captures and formalizes change in the business process.

Rest of the paper is structured as follows. Section II provides the top level framework for the change management approach. Section III discusses an example of a business process and defines the scope of this paper. Section IV presents BPCM ontology which is a main contribution of the paper. Section V concludes our paper.

## II. BUSINESS PROCESS CHANGE MANAGEMENT APPROACH

*Business Process Change Management (BPCM)* needs careful analyses regarding needs of changes and their impacts over the business process performance. We designed six-step methodology for the Business process change management illustrated in Figure 1. Firstly, build the preposition of change in a human understandable language by involving all the stakeholders and participants. Secondly, measure the potential of change, why it is necessary and what are the main positive points to achieve. Mostly, positive points are obvious regarding the proposition, but have some hidden impacts. Thirdly, analyze the side effects over BP and its performance. For this, one has to involve all the participants and stakeholders to gather their opinion and investigate the implication of the changes to be made.

Fourthly formalize and implement the change. This step has many sub tasks as illustrated in Figure 2. First, design the changed BP with some design tool such as Activiti, Eclipse Designer, ARIS, etc. Second, embed the change inside the BP with proposed modifications to have executable BP. Third, record changes in the BPCM ontology (proposed in section iv) to have formal specification of change made inside the BP. Fourth, once we have executable, run the BP to evaluate the changes done and observe its behavior with the modification. If we have several proposed changes, perform these steps for each of the individual change.

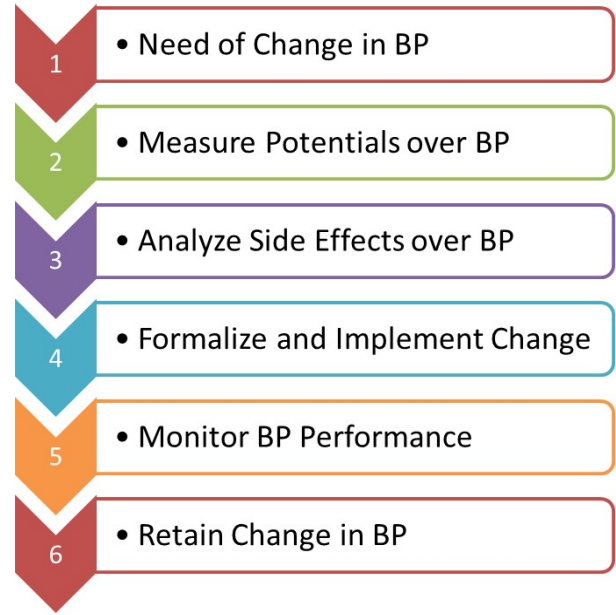

Figure 1.  Business Process Change Management Approach

Once all the changes are done and captured in the formal BPCM ontology, execute the business process and check whether the goal is attained with the modified changes. Lastly, with a positive observation that BP has attained the goals behind the change done, retain the change once it has been made in the BP.

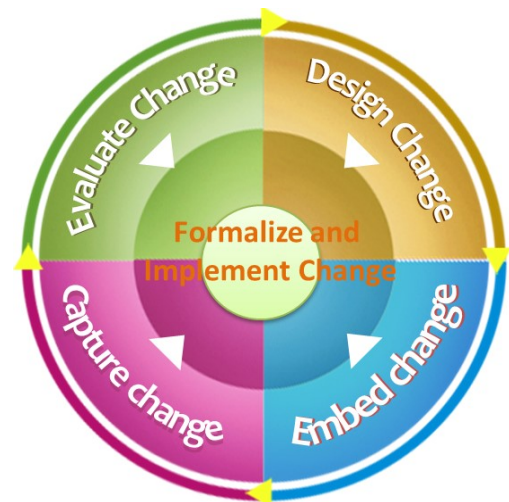

Figure 2.  Steps for Formalizing and Implementing Change

## III. EXAMPLE SCENARIO

In this section, we discuss an example scenario of very small business process that promotes understanding of BPMC ontology and also helps us to define the paper scope.

Consider a *Create Quote* business process illustrated in Figure 3. It has four constructs start, user task, service task and end. A user task is used to model work that needs to be done by a human (or agent). When the user task *'Enter Quotation'* gets execution, it calls the indicated form, displays and gets the user input. After the user submission, a service task *'Register Demand'* is executed that records the input data inside the database. Graphically, a service task looks same for the Java Service Task and Web Service Task, but have different implementation.

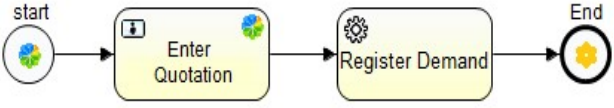

Figure 3. Create Quote Business Process

There can be different types of changes that can occur in these tasks. Let us take first consider a user task. Different changes can be possible such as the *Assignee* (i.e., one who has authorization/ownership to execute) of the user task can be replaced by another person, description of a user task can be changed, due date can be extended, etc. For the service task, changes can be possible such as invocation method can be changed, location or endpoint of a service can be shifted, input/output parameters can be added/deleted, etc. For all these changes, there should be some formal artifact that captures these changes and build a formal change specification. And when there is a need, one can follow the change specification to serve ones purpose. Therefore, we proposed BPCM ontology for the evolving enterprises under collaborative business platform. BPMC ontology captures a very big as per BPMN 2.0 specification, therefore its exhaustive discussion is not possible in this paper. The whole ontology can be downloaded from our personal website [11]. Here, we only discuss changes at user task, and java service task.

## IV. BUSINESS PROCESS CHANGE MANAGEMENT ONTOLOGY (BPCMONT)

Figure 4 illustrates the top level view of *BPCM ontology (BPCMont)*. It captures all the information regarding changes in BPMN constructs, Provenance related to who is going to perform change inside BP and its need/cause/description and Timestamp when the change is implemented. These three top level concepts are explained below.

### A. Provenance Specs

This concept has three sub-concepts. *AgentName* concept captures the name of person or agent who has made the change, *Cause* concept registers the need or cause behind the change, and *Description* concept records information about the change itself in high level language.

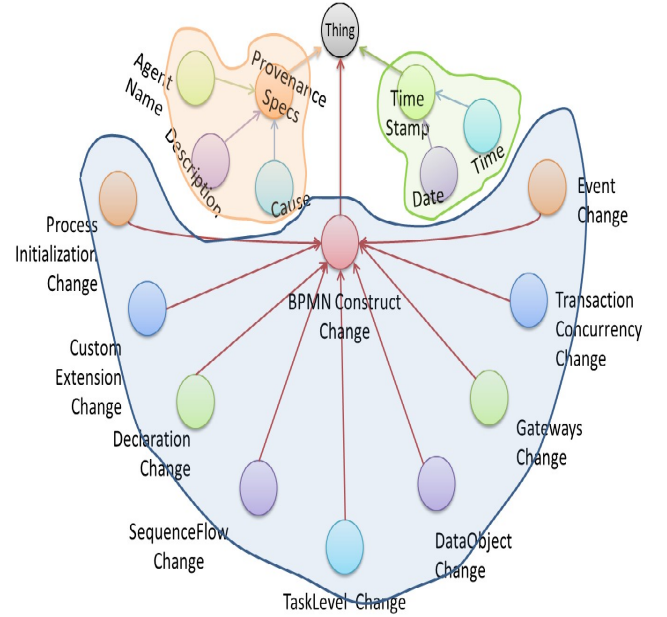

Figure 4. Top level view of BPCM ontology

### B. Timestamp

This concept captures the date and time information when the change has made in *BP*.

### C. BPMN Construct Change

This concept captures the information about which *BPMN construct* has changed, i.e., the change belongs to sequence, task or gateway, etc. It registers all the information regarding particular BPMN constructs such as: *Declaration Change, Process Initialization Change, Sequence Flow Change, Task Level Change, Custom Extension Change, Data Object Change, Gateways Change, Transaction Concurrency Change, and Event Change.*

*Task Level Change.* This concept captures all the information about the change at the task level. For example change within a user task about its name, authorization, addition or deletion, etc. is captured within the UserTask Change concept. Likewise it registers all the information in the particular BPMN construct to which the change has been made.

TaskLevel Change concept comprise of these sub-concepts: *UserTask Change, Java Service Task Change, Web Service Task Change, Script Task Change, Email Task Change; Java Receive Task Change, Business Rule Task Change; Mule Task Change, Manual Task Change, Shell Task Change, Camel Task Change*. Figure 5 shows the class hierarchy of TaskLevel Change concept.

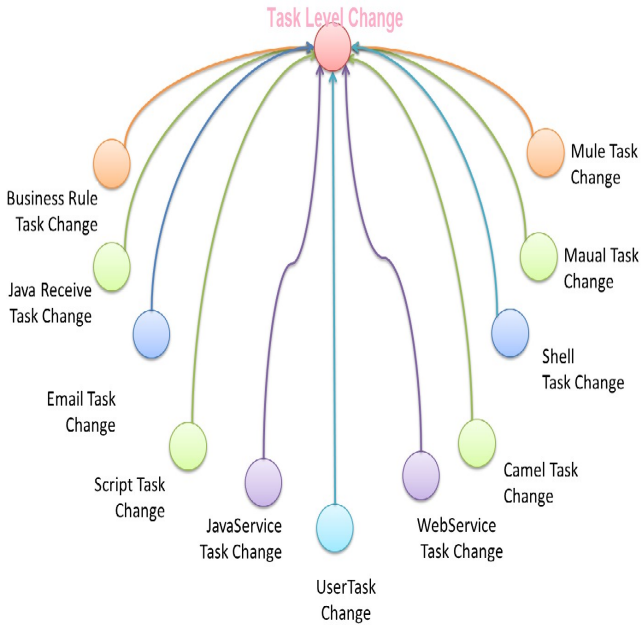

Figure 5. Task Level Change Concept in BPCM ontology

*User Task Change.* UserTask Change concepts captures changes made in the user task. In BP, a user task can be added, deleted, or modified. Modification_in_UserTask Concept records changes regarding its due date, description and/or the user(s) or group(s) assigned to that task. Figure 6 illustrates the hierarchy of UserTask Change concept.

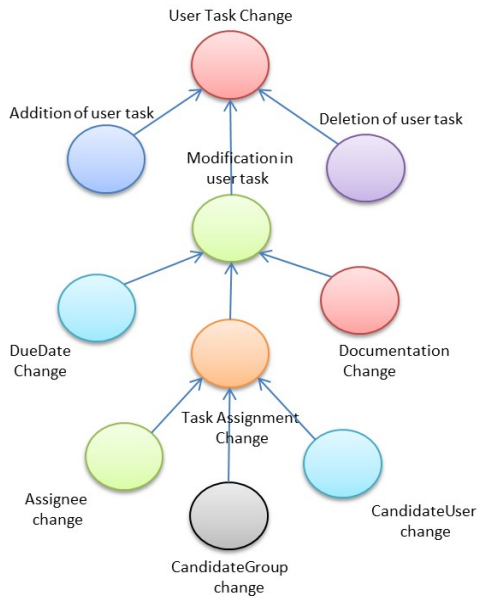

Figure 6. UserTask_Change concept hierarchy in the BPCM ontology

*Java Service Task.* It is used to invoke the web service inside a business process that performs a part of BP transection or workflow. Besides addition, deletion and rename, several modifications can be possible in the java service task.
First type of change can be possible in the method followed to invoke a web service. A web service can be invoked by an intermediate java class, delegation object and method/value expression. Therefore, CallType_Change concept with it corresponding sub-concept records the changes that are made in the java service task. Second type of change can be possible for the input parameters to the service. For the invocation of services, several input parameters can be introduced to inject values into the fields of the delegated classes. This type of change is captured inside the Field_Injection_Change concept. Third type of change can be possible regarding the result returned by the web service which is registered as a ResultVariable_Change concept. Figure 7 depicts the hierarchy of Java Service Task Change concept.

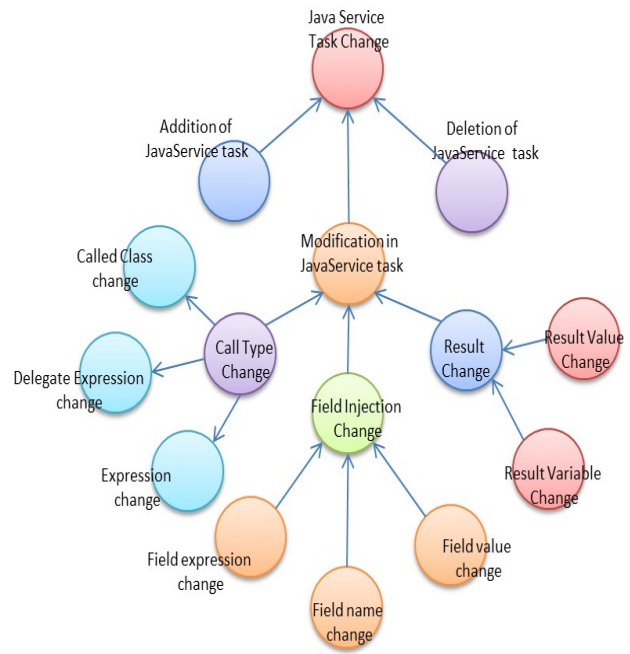

Figure 7. Java Service Task_Change concept hierarchy in the BPCM ontology

V. CONCLUSION AND PERSPECTIVES

Mostly complex business processes are developed and implemented in collaborative environment where many participants are involved to deliver the service and desired functionality. Designing business process is not a one-time exercise. This paper present *Business Process Change Management Ontology (BPCMont)* that serves as a formal specification of all the changes made in the business process in the collaborative environment. We presented different scenarios where BPCMont can be used effectively in the collaborative business platform and presented that it is a significant mile stone towards formalization of changes made in the BPs. we conclude that BPCMont, as a first mile stone, helps enterprises to capture and formalize changes in their business processes which boost their performances by ensuring their ability to change.